\documentclass[empty,a4paper]{llncs}

\usepackage{graphics,graphicx,amssymb,epsfig,balance,algorithmic,subfigure,url,multirow,textcomp,pxfonts,txfonts,float,gensymb}

\newcommand{\keywords}[1]{\par\addvspace\baselineskip
\noindent\keywordname\enspace\ignorespaces#1}

\begin{document}

\mainmatter  

\title{Stratified SIFT Matching for Human Iris Recognition}


%
%
\author{Sambit Bakshi\and Hunny Mehrotra\and Banshidhar Majhi}
%

\institute{Department of Computer Science and Engineering,\\
National Institute of Technology Rourkela\\
Odisha, India\\
sambitbaksi@gmail.com, hunny@nitrkl.ac.in, bmajhi@nitrkl.ac.in\\
http://www.nitrkl.ac.in}

%
%

\maketitle

\begin{abstract}
This paper proposes an efficient three fold stratified SIFT matching for iris recognition. The objective is to filter wrongly paired conventional SIFT matches. In Strata I, the keypoints from gallery and probe iris images are paired using traditional SIFT approach. Due to high image similarity at different regions of iris there may be some impairments. These are detected and filtered by finding gradient of paired keypoints in Strata II. Further, the scaling factor of paired keypoints is used to remove impairments in Strata III. The pairs retained after Strata III are likely to be potential matches for iris recognition. The proposed system performs with an accuracy of 96.08\% and 97.15\% on publicly available CASIAV3 and BATH databases respectively. This marks significant improvement of accuracy and FAR over the existing SIFT matching for iris.
\end{abstract}

\section{Introduction}

\label{sec:intro}
Iris is the sphincter having unique flowery random pattern around the pupil. It is an internal organ with complex unique features that are stable throughout the lifetime of an individual. There has been significant research done in the area of iris recognition using global features \cite{daugman2003pr,iriswavelet,dctiris,christel2002vi}. However, these approaches fail to possess invariance to affine transformations, occlusion and robustness to unconstrained iris images. Thus, there is a stringent requirement to develop iris recognition system suitable for non-cooperative images. Keypoint descriptors are invariant to affine transformation as well as partial occlusion. Scale Invariant Feature Transform (SIFT) is a well known keypoint descriptor for object recognition \cite{dglowe2004}. Due to inherent advantages, SIFT is capable of performing recognition using non-cooperative iris images \cite{regionSift}. In SIFT matching approach, the difference of Gaussian images are used to identify keypoints at varying scale and orientation. The orientation is assigned to each detected keypoint and a window is taken relative to direction of orientation to find the descriptor vector. During recognition, the keypoints are detected from gallery and probe images and matching is performed using nearest neighbour approach. The challenge with conventional SIFT matching when applied to iris recognition is to find texture similarity between same regions of two iris. 
\begin{figure}[h!]
\centering
\subfigure[Strata I: $\eta$ = 98]{
\includegraphics[width=0.40\textwidth]{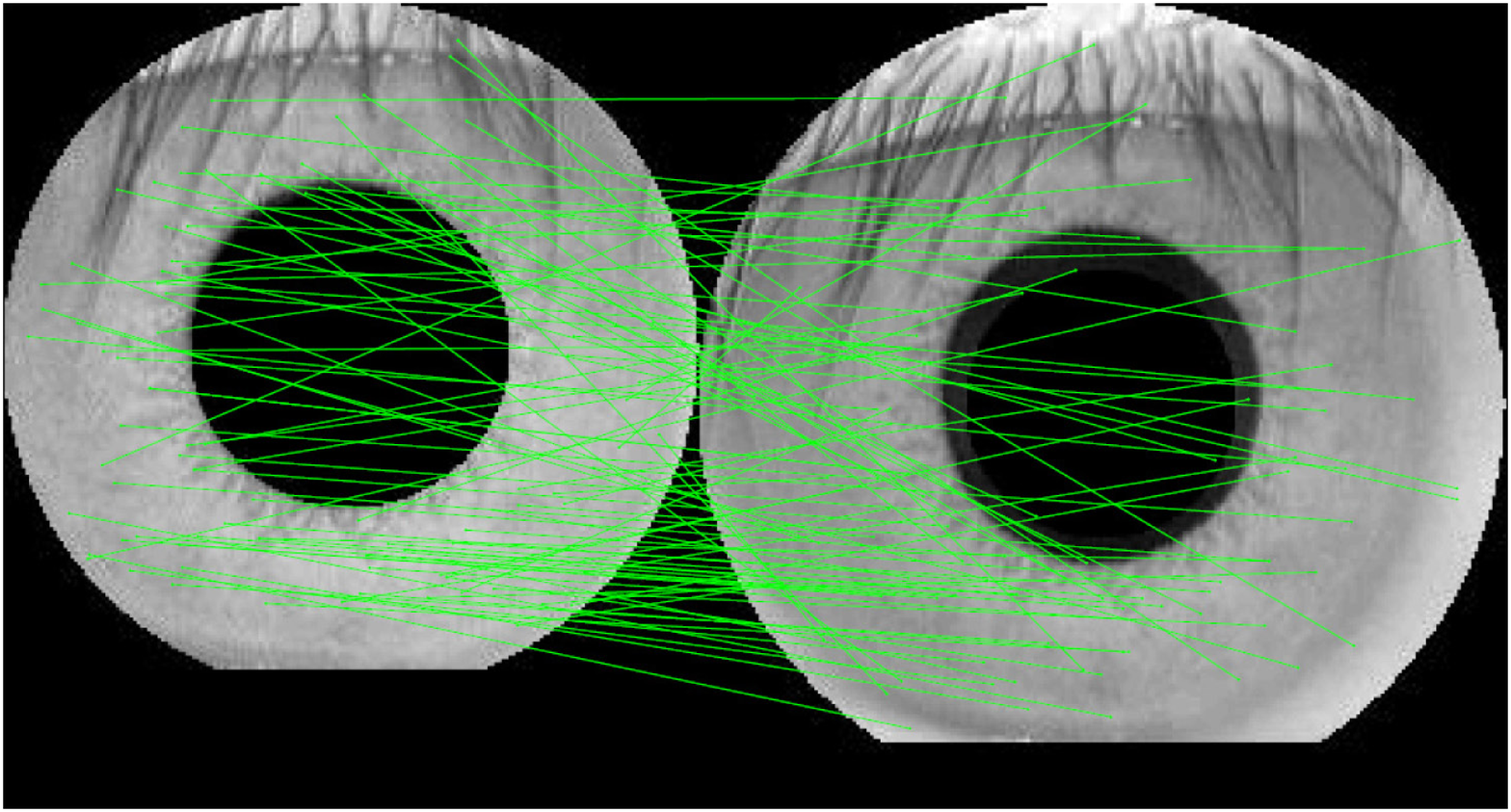}
\label{fig:siftmatch}
}
\subfigure[Strata II: $\eta$ = 65]{
\includegraphics[width=0.40\textwidth]{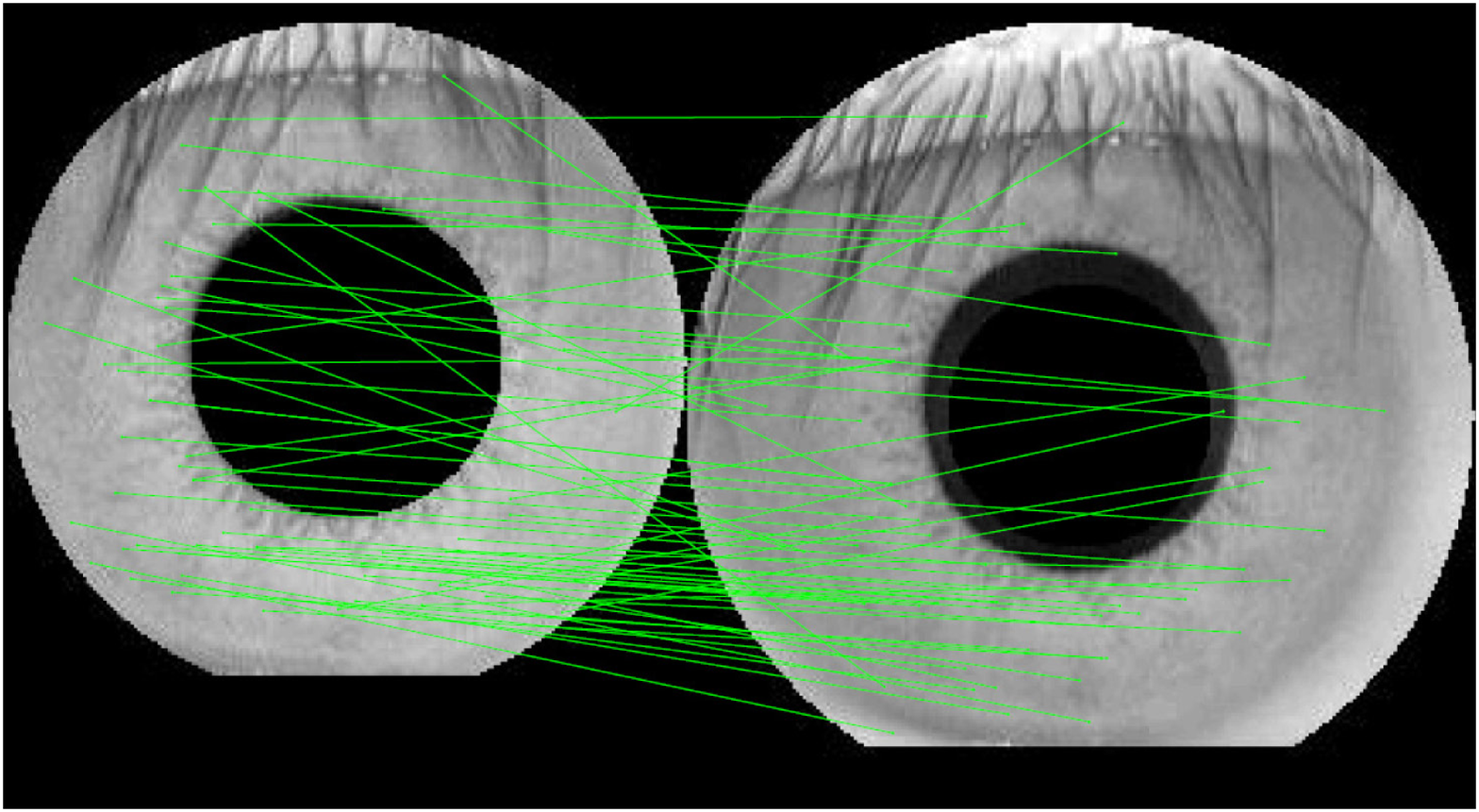}
\label{fig:proposedmatch1}
}
\subfigure[Strata III: $\eta$ = 54]{
\includegraphics[width=0.40\textwidth]{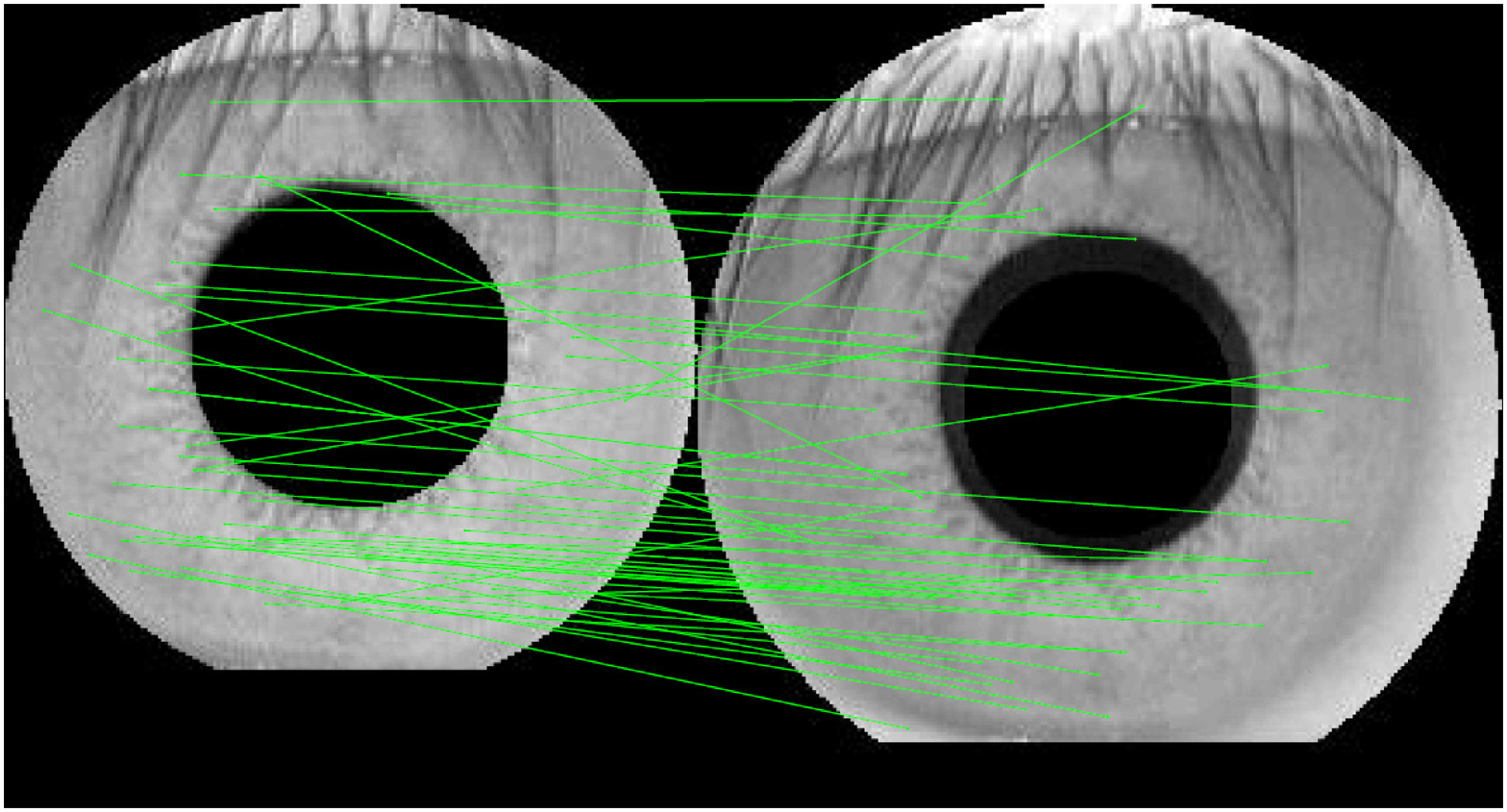}
\label{fig:proposedmatch2}
}
\caption{Matches ($\eta$) at different strata obtained between two instances of same individual taken from CASIAV3}
\end{figure}
SIFT fails as it does not consider spatial information of the keypoints. To make SIFT functional for iris, a novel matching approach has been developed that combines spatial information along with local descriptor of each keypoint. 

The organization of the paper is as follows. Conventional SIFT matching approach for local descriptors is discussed in Section~\ref{sec:consift}. Proposed stratified SIFT matching is explained in detail in Section~\ref{sec:proposedapproach}. The job of filtering is done in two sequencial steps: Gradient based filtering and Scale based filtering. Section~\ref{sec:results} provides the results obtained using proposed approach. Finally, conclusions are given at the end of the paper.

\section{Conventional SIFT Matching}
\label{sec:consift}
The matching algorithm plays a significant role in any biometric system as it acts like a one way gateway through which only genuine matches (if two images are from same subject) will pass and imposter matches (if two images are from different subjects) are blocked. In local feature matching, the total number of paired keypoints is used to find the authenticity of an individual. Let $I$ be the set of all images available in the iris database. For understanding, $I_m$ be a gallery iris image and $I_n$ be a probe iris image where $I_m,I_n \epsilon \ I$. The matching algorithm validates $I_n$ against $I_m$. In the conventional SIFT matching, for each keypoint in $I_m$ the Euclidean distance is found with every keypoint in $I_n$. The nearest neighbour approach pairs the $i^{th}$ keypoint in $I_m$ with $j^{th}$ keypoint in $I_n$, iff the descriptor distance between the two (after multiplying with a threshold) is minimum \cite{dglowe2004}. The two keypoints are paired and removed from the set of keypoints detected from $I_m$ and $I_n$. This process is iterated for the remaining keypoints until any two keypoints can be matched. This approach performs moderately well for unconstrained iris recognition \cite{regionSift}. However, as SIFT determines image similarity using 128-dimensional local features only, hence it may wrongly pair (impairment) some keypoints for iris. Thus, the existing approach is modified using two strata which removes impaired matches using spatial information of the matching keypoints and contributes in achieveing better recognition accuracy.

\section{Stratified SIFT Matching}
\label{sec:proposedapproach}
In the proposed paper an effort is made to improvise the conventional SIFT matching. The pupil and iris circles are assumed to be concentric, hence all localized images  have pixel size $2r\times 2r$, where $r$ is the radius of iris. The pupil center as well as iris centre are located at $(r,r)$. Therefore the localized images do not have transformation due to translation. However, there is a possibility of iris images being transformed due to rotation (tilt of subject's head), scaling (change in camera to eye distance) or both~\cite{regionSift}. The SIFT matching algorithm matches keypoints that have similarity between the local descriptors (as discussed in Section~\ref{sec:consift}) but fails to conform to spatial relationship. The removal (filtering) of impairments by the proposed approach retains only those matches that are more probable to be potential matches. Let $K_m$ be the set of $m_1$ keypoints found in $I_m$ and $K_n$ be the set of $n_1$ keypoints found in $I_n$ by applying SIFT detector. These sets of keypoints are used to comprehend the stratified SIFT matching.

\subsection{Strata I: SIFT Matching}
Let $R$ be the ordered set containing the matches between $K_m$ and $K_n$ by conventional SIFT matching as discussed in Section~\ref{sec:consift}. Hence, $R$ contains only those pairs $(i,j)$ where $i^{th}$ keypoint in $K_m$ is matched with $j^{th}$ keypoint in $K_n$ as shown in Fig.~\ref{fig:siftmatch}. Let $\eta$ be the number of matches found where $\eta\ \epsilon\ [0, \mbox{min}(m_1,n_1)]$. As set $R$ is generated solely on the basis of local descriptor property it may wrongly pair keypoints from different regions of iris. Hence, there is a need to combine spatial information with local descriptor to filter out impaired keypoints as discussed in subsequent strata. 

\subsection{Strata II: Gradient Based Filtering}
\label{ssec:rotation}

\begin{figure}[h!]
\centering
\includegraphics[width=0.60\textwidth]{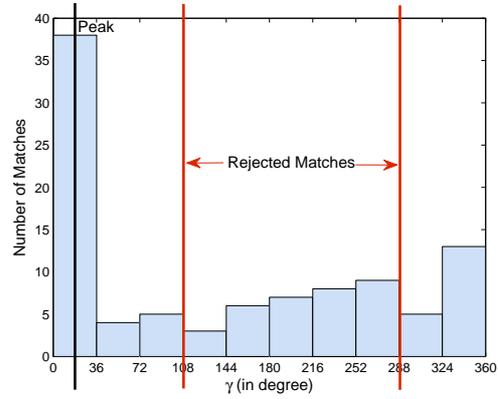}
\caption{Distribution of $\gamma$ for number of matches between two instances of same individual from CASIAV3}
\label{fig:gradhist}
\end{figure}

\begin{figure*}[h!]
\centering
\subfigure[$\gamma = (\phi_j - \theta_i)\ \mbox{mod}\ 360\degree$]{
\includegraphics[width=0.40\textwidth]{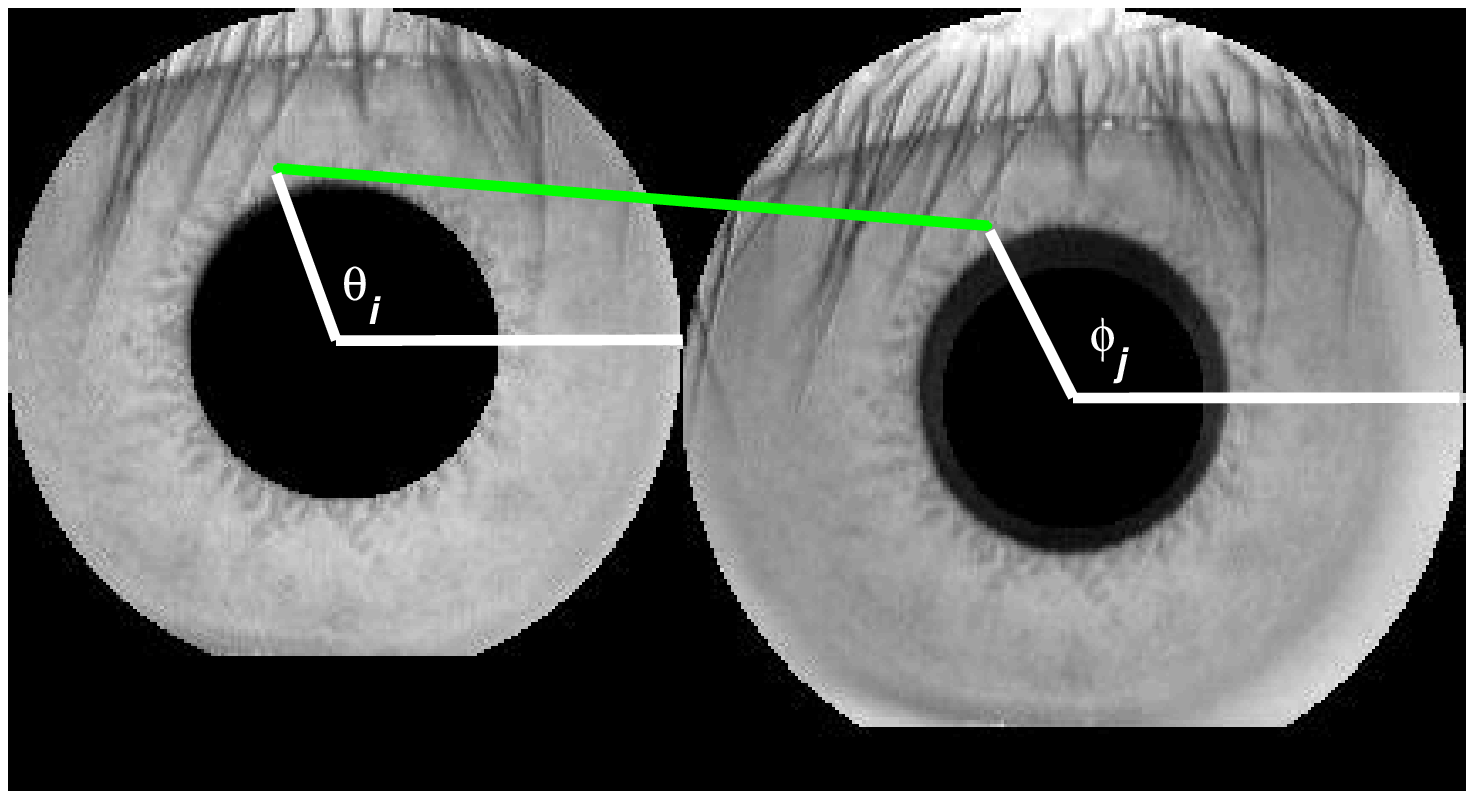}
\label{fig:strata2}
}
\subfigure[$\psi = \frac{d_2}{d_1}$]{
\includegraphics[width=0.40\textwidth]{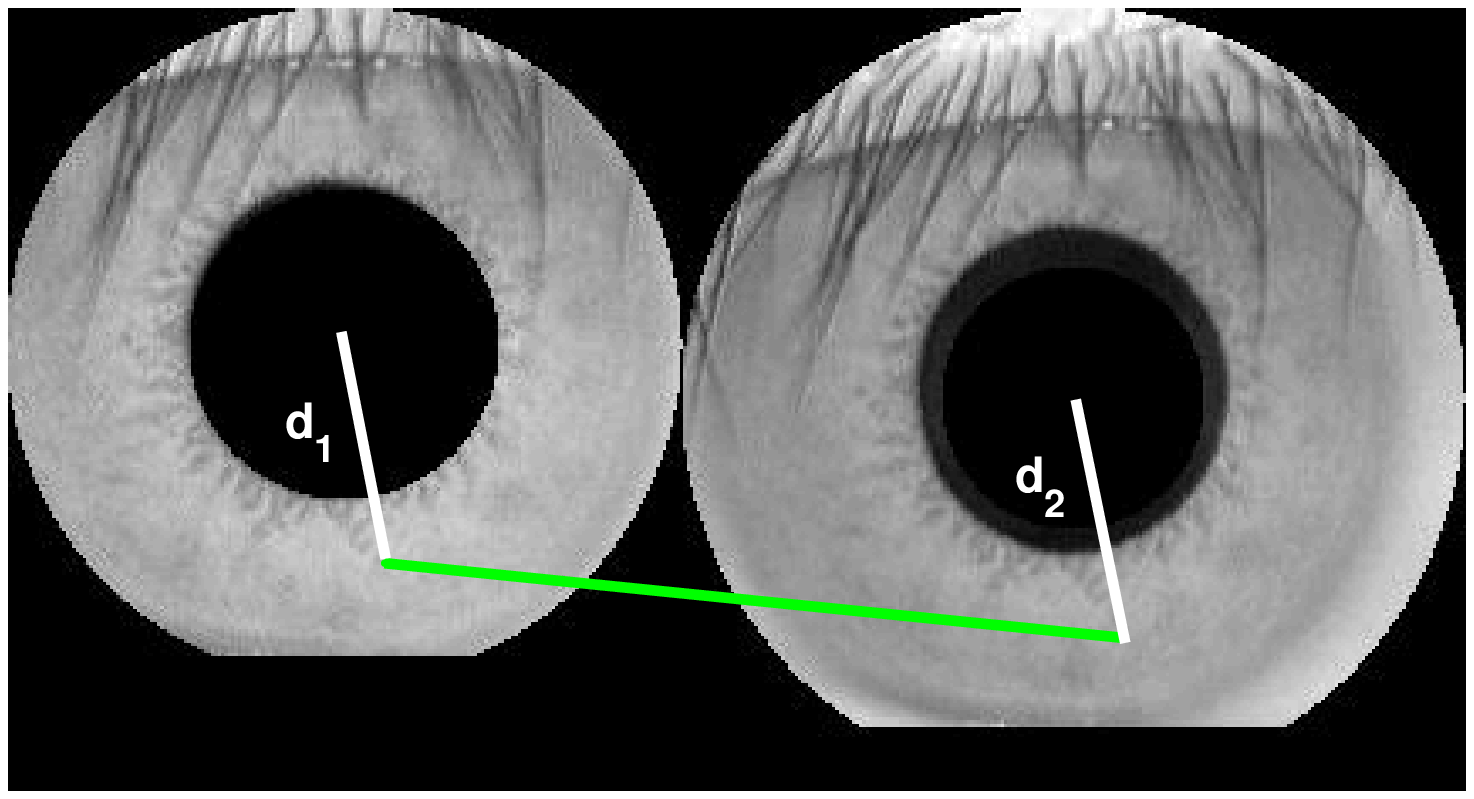}
\label{fig:strata3}
}
\caption{(a) Gradient computation in Strata II, (b) Local scaling factor computation in Strata III}
\end{figure*}

In this strata gradient based filtering is performed to remove impairments from $R$. To compute gradient for each pair of keypoints $(i,j)$ in $R$, the angles are obtained from respective image centers $(r,r)$. Thus, $\theta_i$ is computed from $I_m$ and $\phi_j$ is computed from $I_n$. The angle of rotation for $k^{th}$ pair is calculated as $\gamma_k=(\phi_j-\Theta_i)$ mod~$360\degree$ (depicted in Fig.~\ref{fig:strata2}). Considering SIFT to be completely \emph{flawless} (due to robustness property, no false match is found) and \emph{efficient} (due to property of repeatability, all possible matches are found)~\cite{dglowe2004}; the value of $\gamma_k$ derived should be same $\forall k$. But in practice, SIFT does not give such precise matches. Thus, it is difficult to obtain unique value of $\gamma$ even when $I_m$ and $I_n$ belong to the same subject. Rather a distribution of $\gamma$ is obtained. A histogram is plotted with x-axis comprising bins with a range of values of $\gamma$, and y-axis comprising number of matches falling in a particular bin as shown in Fig.~\ref{fig:gradhist}. The number of bins in the histogram ($nobins$) is subject to implementation issue. In proposed system, $nobins$ is taken as 10. The distribution of $\gamma$ gives a single peak in case the two iris images ($I_m$ and $I_n$) are from the same subject. In contrast, no distinct peak should be found in case the two iris images are from different subjects. There may be error due to discretization of bins, so two adjacent bins of the peak are combined to improve peak density (number of matches). The idea is to find whether the density of the peak exceeds the boundary criteria. It is inferred that a peak is strong if the density exceeds certain higher bound ($hp\%$ of total number of matches).

Likewise peak is weak if the density is less than lower bound ($lp\%$ of total number of matches). If a strong peak is found, an angular range is specified around the peak. Those matches in $R$ for which $\gamma$ are not within the angular range are found to be impaired and removed from $R$ to generate $Rinter$.

For example, as shown in histogram in Fig.~\ref{fig:gradhist}, the peak is found at $0^{th}$ bin which represents gradient value of $0\degree$ to $36\degree$ with a central value of $18\degree$. Hence only those pairs having angular range between $(18\pm 90)$ mod $360\degree$ are retained. Thus, it is evident that $Rinter\subseteq R$ after removing some impairments. Fig.\ref{fig:proposedmatch1} shows paired keypoints in $Rinter$ with considerable reduction of $\eta$. If no strong peak is found it is inferred that all matches in $R$ are faulty, and removed. As a result $Rinter$ becomes empty. 

\subsection{Strata III: Scale Based Filtering}
\label{ssec:scaling}

In this strata further filtering of $Rinter$ is performed on the basis of global and local scaling factor between the gallery and probe images. The global scaling factor ($sf$) between two images is defined as ratio of probe iris radius $(r_{n})$ to gallery iris radius $(r_m)$. A scale range with certain tolerance around $sf$ is empirically taken to handle aliasing artifact. From implementation perspective, the scale range is taken as ($sf \pm 0.2$). In ideal case, if gallery and probe belong to same individual the scaling factor between all paired keypoints should be unique. However, this does not hold good in practical scenarios. 

During filtering, for each element in $Rinter$, two Euclidean distances are calculated- (a) $d_1$: distance of $i^{th}$ keypoint of $I_m$ from its center and (b) $d_2$: distance of $j^{th}$ keypoint of $I_n$ from its center. Local scaling factor ($\psi$) for each element of $Rinter$ is calculated as $\psi = d_2/d_1$ (shown in Fig.~\ref{fig:strata3}). Matches having $\psi$ within scale range discussed above qualifies to be potential and stored in $Rnew$, else are labeled as faulty and filtered. Fig.\ref{fig:proposedmatch2} shows paired keypoints in $Rnew$ after further reduction of $\eta$.

\section{EXPERIMENTAL RESULTS}
\label{sec:results}

\begin{figure*}[h!]
\centering
\subfigure[]{
\includegraphics[width=0.48\textwidth]{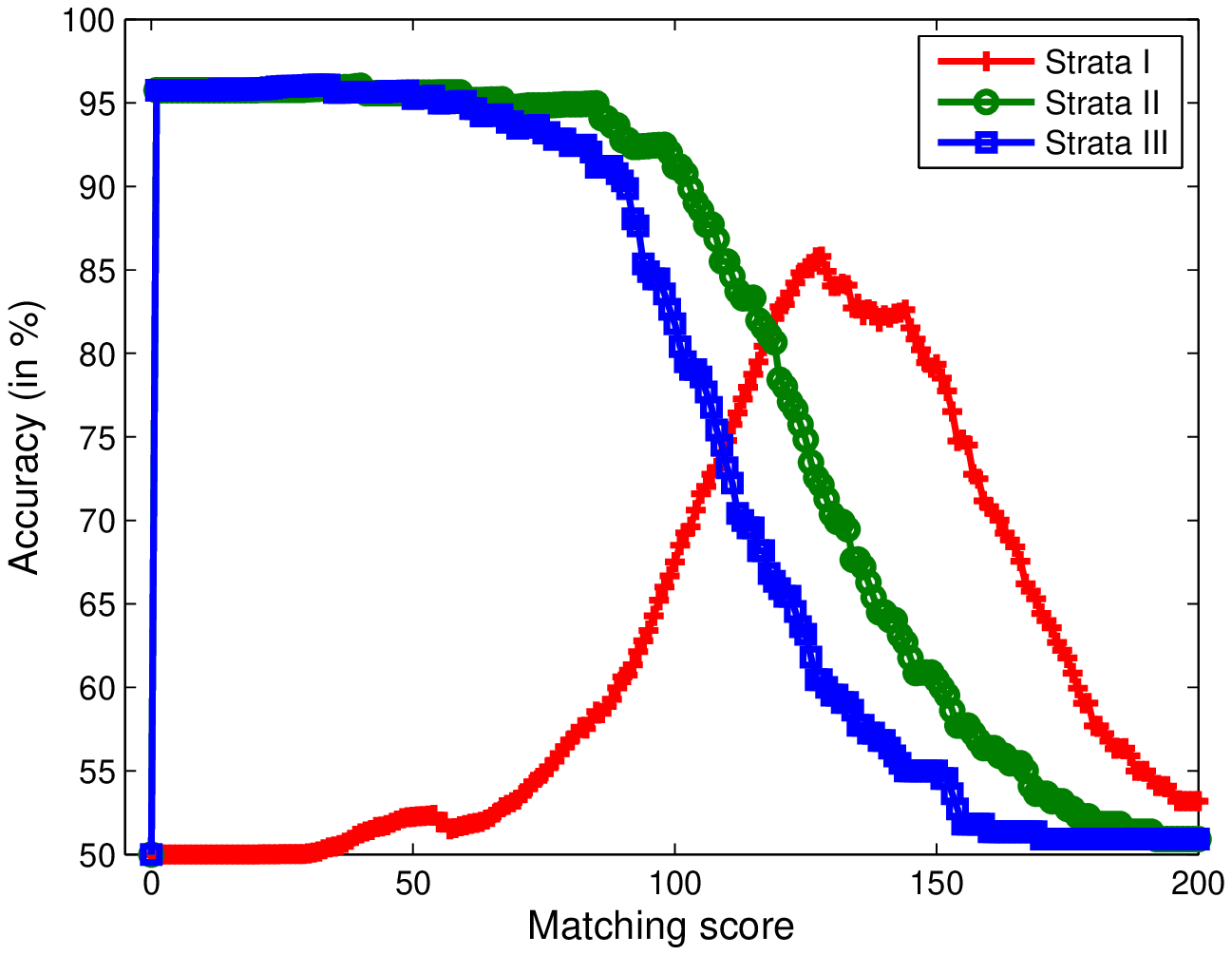}
\label{fig:casiaAcc}
}
\subfigure[]{
\includegraphics[width=0.48\textwidth]{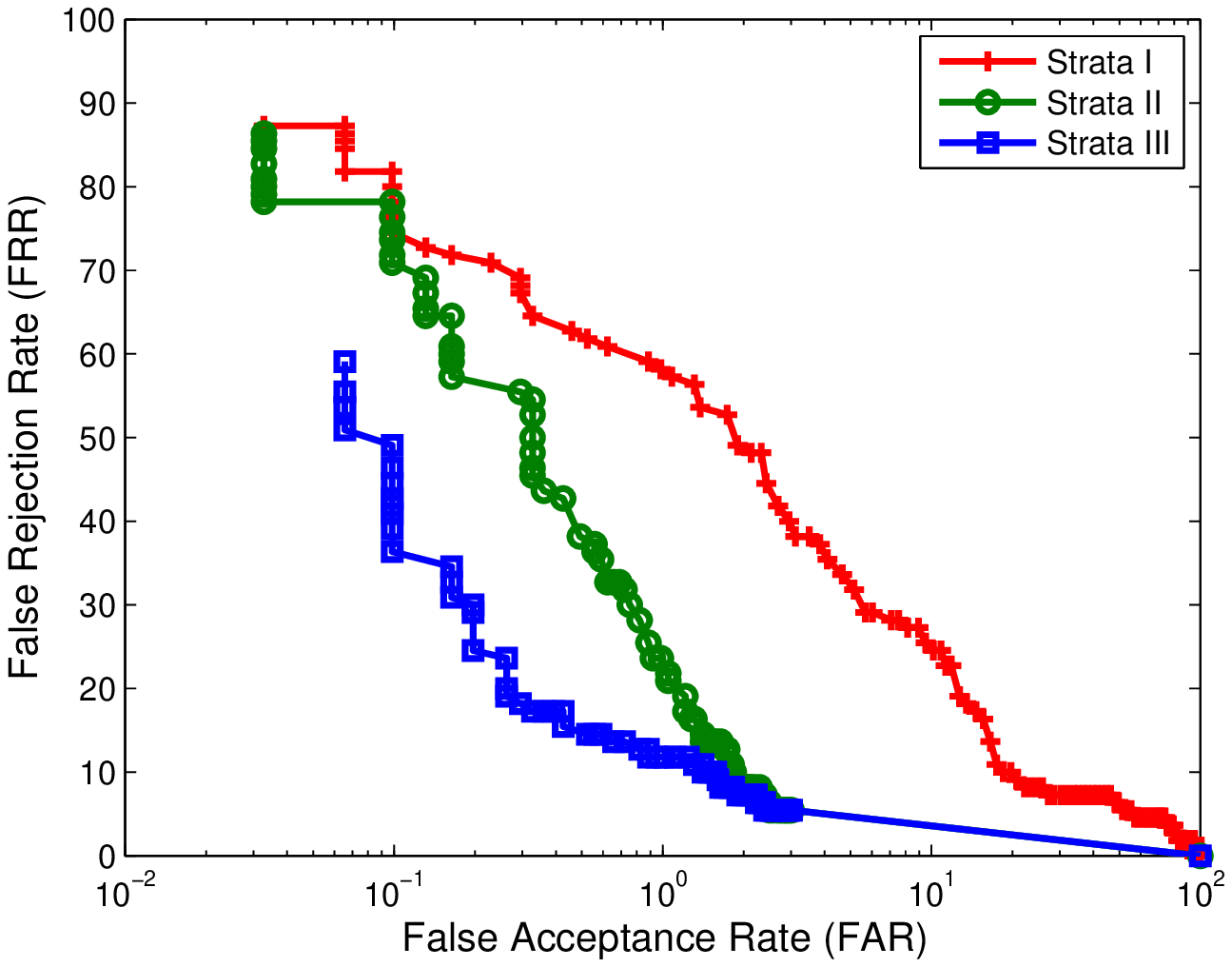}
\label{fig:casiaRoc}
}
\caption{(a) Accuracy curve for three strata, (b) ROC curve for three strata on CASIAV3}
\end{figure*}

The proposed stratified SIFT matching is tested on publicly available BATH~\cite{bath} and CASIAV3 (CASIAV3)~\cite{casiav3} databases. Database available from BATH University includes images from 50 subjects (20 images per subject from both the eyes). CASIAV3 (CASIAV3) comprises 249 subjects with total of 2655 images from both the eyes. 
The experiments are carried out on 2.13GHz Intel(R) CPU using Matlab. To validate the system performance some standard error measures~\cite{Jain:2007:HB:1324787} are used\footnote{FAR: False Acceptance Rate, FRR: False Rejection Rate, ACC: Accuracy, d': d-prime value}. The results are carried out in three different strata as given in Table~\ref{tab:accuracy}. In \emph{Strata I}, the two iris images are matched using conventional SIFT approach. This approach performs with an accuracy of 85.81\% on CASIAV3 database.
\begin{table*}[h!]
\centering
\caption{Performance measures for stratified SIFT matching}
\vspace{0.1in}
\begin{tabular}{|l|c|c|c|c|c|c|c|c|}
\hline
{\bf Databases $\rightarrow$}
 & \multicolumn{4}{|c|}{\bf CASIAV3} & \multicolumn{4}{|c|}{\bf BATH}\\
\hline
{\bf Approach $\downarrow$}&{\bf FAR} & {\bf FRR} & {\bf ACC} & {\bf d'} & {\bf FAR} & {\bf FRR} & {\bf ACC} & {\bf d'} \\
\hline
{\bf Strata I (Conventional SIFT)} & 17.48 & 10.91 & 85.81 & 1.20 & 1.57 & 4.35 & 97.04 & 2.73\\\hline
{\bf Strata II (Gradient based Filtering)} & 2.49 & 5.45 & 96.03 & 2.46 & 0.97 & 6.09 & 96.47 & 2.81\\\hline
{\bf Strata III (Scale based Filtering)} & 2.39 & 5.45 & 96.08 & 2.20 & 1.34 & 4.35 & 97.15 & 2.90\\\hline
\end{tabular}
\label{tab:accuracy}
\end{table*}
Likewise, for BATH database an accuracy of 97.04\% is obtained. To improve the performance of the system, the objective of the proposed research is to reduce false acceptances. In \emph{Strata II}, the impairments are removed using gradient filtering which significantly increases the seperability measure between false and genuine matches as indicated by d' values given in the Table~\ref{tab:accuracy}. Further improvement in separability and accuracy are brought by  scale filtering in \emph{Strata III}. The accuracy values are plotted against change in number of matches as shown in Fig.~\ref{fig:casiaAcc}.

\begin{figure}[h!]
\centering
\includegraphics[width=0.65\textwidth]{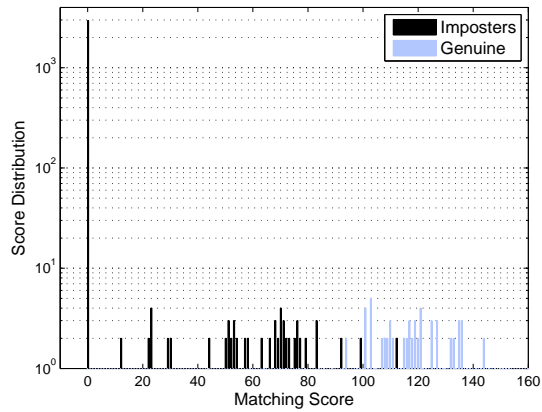}
\caption{Final histogram of scores on CASIAV3}
\label{fig:casiaHist}
\end{figure}

 The Receiver Operating Characteristic (ROC) curves~\cite{Jain:2007:HB:1324787} for three different strata are shown in Fig.~\ref{fig:casiaRoc}. The distribution of genuine and imposter scores after Strata III is shown in Fig.~\ref{fig:casiaHist}. All graphical results are obtained for CASIAV3 and similar observations are made for BATH database.

\section{CONCLUSIONS}
\label{sec:conclusions}
In this paper, a novel stratified SIFT matching technique is proposed that improvises conventional SIFT by removing wrong pairs. This approach provides boost in accuracy due to considerable reduction in FAR. The FAR is reduced by 15.09\% and 0.23\% for CASIAV3 and BATH respectively. From the results it has been observed that the proposed algorithm is completely flawless, i.e., matches removed are guaranteed to be wrong matches whereas it is not completely efficient, i.e., all impairments by SIFT are not guaranteed to be filtered. However, the gain in accuracy is substantial which marks its applicability for unconstrained iris recognition.

\end{document}